\documentclass[journal]{IEEEtran}
\usepackage{amsmath}
\usepackage{graphicx}
\usepackage[hidelinks]{hyperref}
\usepackage{amsthm}
\usepackage{multirow}
\usepackage{booktabs}
\usepackage{algorithm}
\usepackage{algorithmic}
\usepackage{color}
\usepackage{amsfonts}
\usepackage{ulem}
\usepackage{bm}
\usepackage{bbm}
\usepackage{subfigure}
\newtheorem{theorem}{Theorem}

\newtheorem{definition}{Definition}
\newtheorem{assumption}{Assumption}

\ifCLASSOPTIONcompsoc
  \usepackage[nocompress]{cite}
\else
  \usepackage{cite}
\fi
\ifCLASSINFOpdf
\else
\fi
\begin{document}
\renewcommand\arraystretch{1.3}
%
\title{Time-Series Domain Adaptation via Sparse Associative Structure Alignment: Learning \\ Invariance and Variance}

%

%
%

\author{Zijian Li, Ruichu~Cai,~\IEEEmembership{Senior Member,~IEEE,}  Jiawei Chen, Yuguang Yan, Wei Chen, Keli Zhang, Junjian Ye
\IEEEcompsocitemizethanks{\IEEEcompsocthanksitem Zijian Li, Jiawei Chen, Yuguang Yan are with the School of Computing, Guangdong University of Technology, Guangzhou China, 510006.
E-mail: \{leizigin,  chenjiawei952\}@gmail.com, ygyan@gdut.edu.cn
\IEEEcompsocthanksitem Ruichu Cai, Wei Chen are with the School of Computer Science, Guangdong University of Technology, Guangzhou, China, 510006 and Peng Cheng Laboratory, Shenzhen, China, 518066.
E-mail: \{cairuichu, chenweidelight\}@gmail.com
\IEEEcompsocthanksitem Keli Zhang, Junjian Ye is with Huawei Noah's Ark Lab, Huanwei, Shenzhen, China, 518116
E-mail: \{zhangkeli1, yejunjian\}@huawei.com
}
\thanks{Manuscript received XX; revised XX; accepted XX. Date of publication XX XX, 2019; date of current version XX XX, 2019. This research was supported in part by National Key R\&D Program of China (2021ZD0111501), National Science Fund for Excellent Young Scholars (6212200101) and Natural Science Foundation of China (61876043, 61976052). Wei Chen was supported by China Postdoctoral Science Foundation (2021M690734). (*Ruichu Cai is the Corresponding author.)
}
}

%
%

\markboth{}%
{Ruichu Cai \MakeLowercase{\textit{et al.}}: Time-Series Domain Adaptation via Sparse Associative Structure Alignment: Learning Invariance and Variance}
%



\IEEEtitleabstractindextext{%
\begin{abstract}
\textcolor{black}{Domain adaptation on time-series data is often encountered in the industry but received limited attention in academia}. Most of the existing domain adaptation methods for time-series data \textcolor{black}{borrow the ideas from} the existing methods for non-time series data to extract the domain-invariant representation. 
However, two peculiar difficulties to time-series data have not been solved. 1) \textcolor{black}{It is not a trivial task to model the domain-invariant \textcolor{black}{and} complex dependence among different timestamps.} 2) The domain-variant information is important but how to leverage them is almost underexploited.
Fortunately, the stableness of causal structures among different domains inspires us to explore the structures behind the time-series data. Based on this inspiration, \textcolor{black}{we investigate the domain-invariant unweighted sparse associative structures and the domain-variant strengths of the structures.} 
To achieve this, we propose \textbf{S}parse \textbf{A}ssociative structure alignment by learning \textbf{I}nvariance and \textbf{V}ariance (\textcolor{black}{SASA-IV} in short), a model that simultaneously aligns the invariant unweighted spare associative structures and considers the variant information for time-series unsupervised domain adaptation.
\textcolor{black}{Technologically, we extract the domain-invariant unweighted sparse associative structures with a unidirectional alignment restriction and embed the domain-variant strengths via a well-designed autoregressive module.}
Experimental results not only \textcolor{black}{testify that our model yields state-of-the-art performance} on three real-world datasets but also provide some insightful discoveries on the knowledge transfer.
\end{abstract}

}

\maketitle
\IEEEdisplaynontitleabstractindextext
\IEEEpeerreviewmaketitle

\section{Introduction}\label{sec:intro}
Unsupervised domain adaptation (UDA) \cite{ganin2015unsupervised, tzeng2014deep, pan2009survey, long2016unsupervised} is proposed to address the problem named ``\textit{domain shift}'' \cite{8337102}, \textcolor{black}{in which the source distribution and the target distribution are different}. 
Because of the great success in the non-time series data, many researchers have extended the existing works for non-time series data to the scenario of time-series data. The existing methods \cite{da2020remaining,DBLP:conf/iclr/PurushothamCNL17,ijcai2021-378} for time-series UDA usually combine the neural architectures for time-series data like recurrent neural networks  \cite{mikolov2010recurrent,chung2015recurrent} and adversarial learning methodology like gradient reversal layer (GRL) \cite{ganin2015unsupervised} to extract the domain-invariant representation. Recently, Liu et.al \cite{ijcai2021-378} use the Fourier spectral theory and propose the adversarial spectral kernel matching method to extract the domain-invariant information for time-series data.

\begin{figure}
	\centering
	\includegraphics[width=\columnwidth]{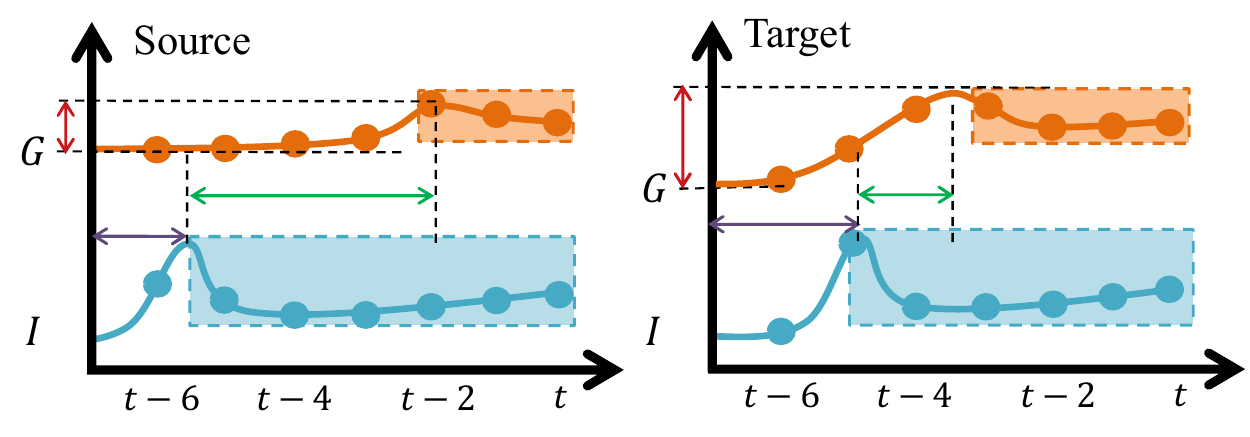}
	\caption{The illustration of the relationships between ``Growth Rate (G)$\downarrow$'' and ``Irrigation Volume (I)$\downarrow$'', the decrease of G results in the decrease of I. The different lengths of red double-head arrows denote different value ranges. The different lengths of blue double-head arrows denote different offsets. The different lengths of green double-head arrows denote different response times. Different response time means different time lags. In growth rate prediction scenario, the conditional distribution $P(G_t|G_{1:t-1}, I_{t-1})$ are influenced by value ranges offsets and time lags. (\textit{Best view in color.})}
	\label{fig:motivation1}
\end{figure}

\begin{figure}[t]
	\centering
	\includegraphics[width=\columnwidth,height=1.0\columnwidth]{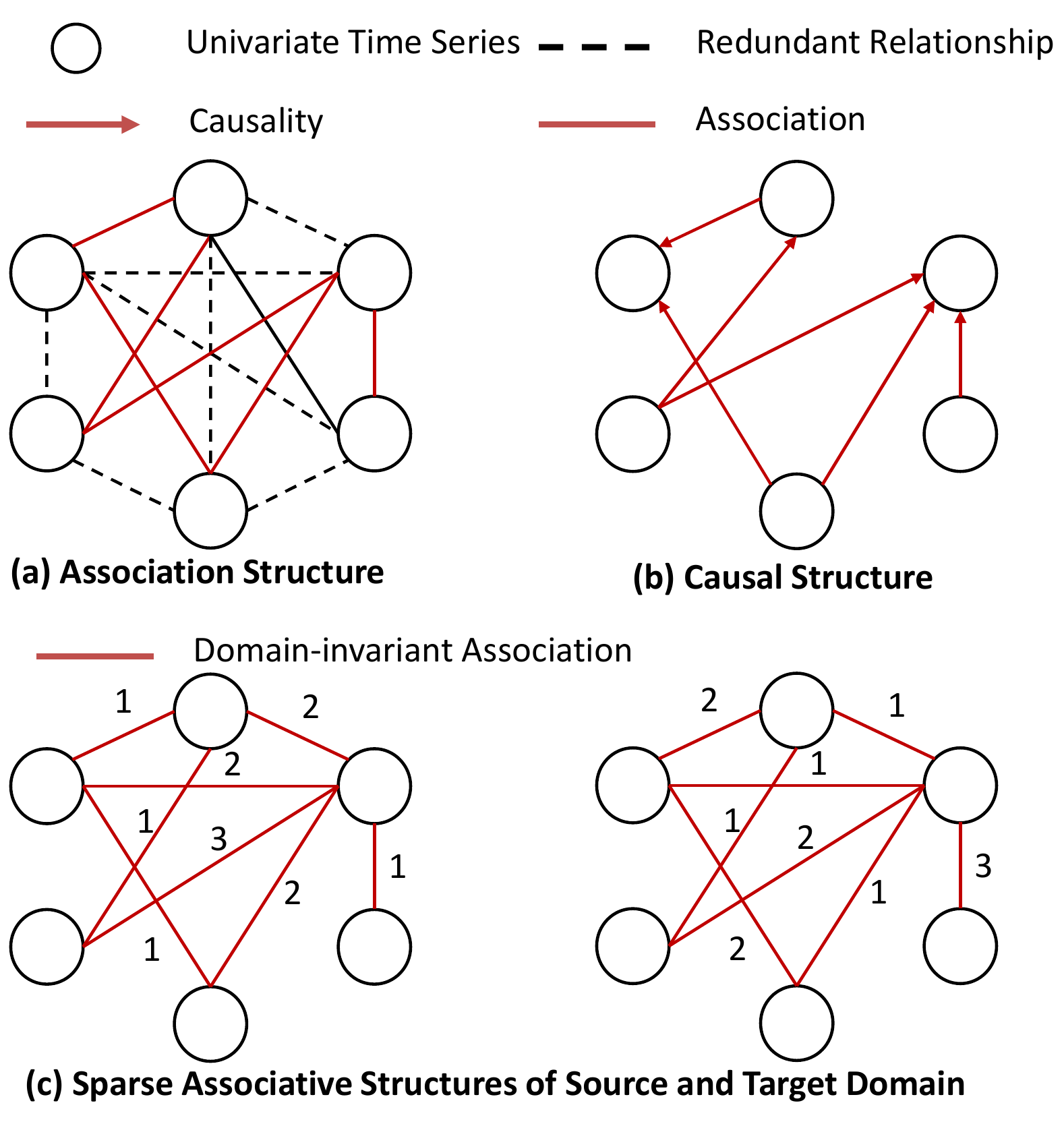}
	\caption{The illustration of various structures among six time series. (a) The existing methods take all the relationships into account and lead to redundancy. (b) The causal structure of variables. (c) Inspired by the stability of the causal mechanism, our method considers both the domain-invariant sparse associative structure and the domain-variant strengths.}
	\label{fig:motivation2}
\end{figure}

Though taking the first step, two core challenges of time-series domain adaptation have not been addressed: \textcolor{black}{(1) How to extract the domain-invariant information for time-series data? (2) How to use the domain-variant factors for model prediction for time-series data?} 
\textcolor{black}{As for the first question, the existing methods assume that the domain-invariant information can be extracted via a single feature extractor, but this assumption is hard to be satisfied in time-series data because these feature extractors are hard to capture how the variables relate to others and even the first-order Markov dependence between any two timestamps results in the variance of conditional distributions of different domains, i.e., $P_S(y_t|\phi(\bm{x}_1,\cdots, \bm{x}_t)) \neq P_T(y_t|\phi(\bm{x}_1,\cdots, \bm{x}_t))$.}
As shown in Figure \ref{fig:motivation1}, even small discrepancies between the source and the target domains may lead to sharp changes in conditional distributions. As for the second question, domain-variant information \textcolor{black}{like the strength of the association} is rarely considered but plays an important role in model prediction. For example, the influence degree between ''Irrigation Volume`` and ''Growth Rate`` varies with different plants, which should be taken into consideration in plant growth rate prediction. 
In summary, the existing methods, which essentially consider both the associations and the redundant relationships like Figure \ref{fig:motivation2} (a), \textcolor{black}{neither capture the domain-invariant sparse association nor leverage the domain-variant strength of association.}

Fortunately, \textcolor{black}{as shown in} Figure \ref{fig:motivation2}(b), the causal mechanisms among different domains are stable (e.g., Irrigation Volume (I) and the Groth Rate (G) shown in Figure \ref{fig:motivation1}), which inspires us to explore the structures behind the time-series data. Excitingly, we find that considering the sparse associative structures can simultaneously bypass the difficulty of causal discovery and address the aforementioned two core challenges in time-series domain adaptation. 
Figure \ref{fig:motivation2}(c) illustrates that the unweighted sparse associative structures from different domains usually coincide with the causal structures and are invariant, while the strengths of the associative structures usually vary with different domains. Therefore, we can solve the time-series domain adaptation problem under the framework of sparse associative structures.



Following the aforementioned intuition, we propose the \textbf{S}parse \textbf{A}ssociative \textbf{S}tructure \textbf{A}lignment by Learning \textbf{I}nvariance and \textbf{V}ariance (\textcolor{black}{SASA-IV} in short) for time-series unsupervised domain adaptation. Technologically, we first propose the adaptive segment summarization to mitigate the obstacle of offset. Second, we extract the unweighted sparse associative structures with the help of intra-variables and the inter-variables attention mechanisms. Third, we encode the domain-variant strengths with the help of a well-designed autoregressive module. Finally, we propose the unidirectional alignment restriction to guarantee the correct transformation direction. Moreover, we also provide theoretical analysis for the proposed methods, in which the generalization risk on the target domain depends on the heuristic structural generation distance. Extensive experimental studies demonstrate that the proposed \textbf{SASA-IV} approach outperforms the several state-of-the-art time-series UDA methods on three real-world datasets.

This paper involves a substantial extension to its conference version \cite{Cai_Chen_Li_Chen_Zhang_Ye_Li_Yang_Zhang_2021}, the extra contributions can be shown as follows:
\begin{itemize}
    \item Compared with the conference version, we discuss what the invariant factor is and what the variant factor is in time-series unsupervised domain adaptation, which provide the more novel insight for time-series unsupervised domain adaptation.
    \item
    We improve the previous SASA model in the following three folds. First, we align the domain-invariant unweighted sparse associative structures for knowledge transfer; Second, we propose the unidirectional alignment restriction to guarantee for correct transformation; Third, we encode the domain-variant strengths with the help of well-designed autoregressive autoencoder.
    \item As for the theoretical analysis, we first propose a novel distribution metric for time-series data then provide a generalization bound for time-series unsupervised domain adaptation for the proposed methods, where the generalization risk on the target domain depends on the risk on the source domain and the distribution metric. 
    \item We compare our method with the latest state-of-the-art on 3 real-world datasets and validate the effectiveness with a series of ablation experiments. Moreover, we provide some insightful visualization results, showing what knowledge can be transferred.
\end{itemize}

The rest of the paper is organized as follows. Section \ref{sec:related} reviews existing studies on domain adaptation for non-time series data and time-series data.  \textcolor{black}{In section \ref{sec:model}, we first provide the problem definition of unsupervised domain adaptation for time-series data. Then we elaborate on the details of the proposed SASA-IV. }We also provide the theoretical analysis in section \ref{sec:theore}. 
Section \ref{sec:exp} presents the experiment results on three real-world datasets, including ablation analysis and the visualization. Section \ref{sec:con} concludes the paper.

\section{Related Works}\label{sec:related}
In this section, we first review the existing techniques about unsupervised domain adaptation for non-time series and time-series data, then we review the works about time-series relational reasoning.

\subsection{Unsupervised Domain Adaptation on Non-Time Series Data.}
To handle the ``\textit{domain shift}'' issue between the source and the target domains, unsupervised domain adaptation has been proposed and applied in various fields \cite{wang2019domain,mahmood2018unsupervised,ramponi2020neural, glorot2011domain,zhang2017curriculum,zou2018unsupervised,hao2021semi}. Most of unsupervised domain adaptation methods follow the covariate shift assumption and aim to extract the domain-invariant representation \cite{6762929}. They can be categorized into the \textit{Maximum Mean Discrepancy based} methods and \textit{adversarial training based} methods. Recently, motivated by the stableness of causality \cite{scholkopf2021toward}, Zhang et.al \cite{cai2019learning, zhang2020domain, zhang2013domain,stojanov2021domain} consider other more challenging scenarios and model how the data distribution changes across different domains.

\subsubsection{Maximum Mean Discrepancy based methods.} The Maximum Mean Discrepancy (MMD) based \cite{chen2021representation,long2018transferable, 8334809} methods \textcolor{black}{used the maximum mean discrepancy to measure and reduce the distance of extracted feature.} Tzeng et.al \cite{tzeng2014deep} introduce the adaptation layer with an additional domain confuse loss to learn the domain-invariant representation. Long et.al \cite{long2015learning} propose the deep adaptation networks to extract the domain-invariant representation with the help of multiple kernel maximum mean discrepancy. Assuming that the source and target classifiers differ by a small residual function, Long et.al \cite{long2016unsupervised} further propose the residual transfer network to explicitly learn the residual function with reference to the target classifier. Yan et.al \cite{Yan_2017_CVPR} consider the changes of class prior distributions and propose the weighed MMD, where the class-specific auxiliary weights are brought into the original MMD.

\subsubsection{Adversarial training based methods.} The adversarial training based methods borrow the ideas of generative adversarial networks \cite{goodfellow2014generative} and extract the domain-invariant representation with the help of the domain classifier. Considering that the domain-invariant representation should be similar, Ganin et.al \cite{ganin2015unsupervised} employ the gradient reversal layer to address the UDA problem. Aiming to minimize the intra-class feature distance, Xie et.al \cite{xie2018learning} propose the moving average centroid alignment method by combining the adversarial training and pseudo label technique. To design a more effective invariant feature space, Gu et.al \cite{gu2020spherical} raise the adversarial domain adaptation method that is defined in the spherical feature space. Recently, Long et.al \cite{zhang2019bridging} introduce the margin disparity discrepancy as a new measurement to generalization bound and implement it with the help of gradient reversal layer.

\subsubsection{Causality based methods.} Since the covariate shift assumption might be not satisfied, Zhang et.al \cite{zhang2013domain} propose the target shift, conditional shift and generalized target shift. Based on the causal generation process, Cai et.al \cite{cai2019learning} propose the disentangled semantic representation framework, in which the semantic information and the domain information are disentangled with the help of the variational autoencoders (VAE) \cite{kingma2013auto}. Ren et.al \cite{8501930} study the generalized conditional domain adaptation problem and propose the propose transforming the class conditional probability matching to the marginal probability matching. Considering the domain adaptation as a graphical inference problem, Zhang et.al \cite{zhang2020domain} model how the joint distribution changes by discovering the causal structures and inferring the label under the causal structures. Recently, Petar and Li et.al \cite{stojanov2021domain} find that domain-invariant representation can not be extracted with the help of a single encoder when the support overlap exists, so they take the domain-specific information into account and propose the domain-specific adversarial network under the causal generation process. 

In this paper, we are also inspired by the stable causal structures assumption. Since discovering the causal structures behind the data is another challenging task that is usually hindered by other factors like hidden confounders, we relax the stable causal structures assumption to stable sparse associative structure assumption and apply it into the time-series UDA problem.

\subsection{Unsupervised Domain Adaptation on Time-series Data.} Time series is one of the most familiar data that can be found in many applications. Time-series domain adaptation is an important problem but desiderates more concern. Recently, more and more attention has been paid to the time-series domain adaptation. Da Costa et al. \cite{da2020remaining} straightforwardly employ the feature extractors that are designed for time-series data like RNN \cite{mikolov2010recurrent} and VRNN \cite{chung2015recurrent} and reuse the conventional unsupervised adaptation frameworks \cite{ganin2015unsupervised}. Wilson et.al \cite{wilson2020multi} propose the CoDATS that leverages the weak supervision in the form of target-domain label distribution. Ragab et.al \cite{ragab2021self} propose the self-supervised autoregressive domain adaptation framework for time-series data which introduces the autoregressive model to model the temporal dependency. Recently, considering that the low-order and local statistics have limited expression for time-series distribution, Liu et.al \cite{ijcai2021-378} propose the adversarial spectral kernel matching method with the help of Fourier transform.

Based on the previous sparse associative structure alignment model (SASA) \cite{Cai_Chen_Li_Chen_Zhang_Ye_Li_Yang_Zhang_2021}, we make a substantial extension and propose SASA-IV. In this paper, we simultaneously consider the invariant unweighted sparse associative structures and the variant strength information. 

\subsection{Time-series Relational Reasoning}
The proposed method also relates to the time-series relational reasoning problem. Aiming to mine the generalized and explainable representation between entities and their properties, time-series relational reasoning \cite{kemp2008discovery} aims to explore the inter-sample relation and intra-temporal relation of time-series data to learn the underlying the structures. In the past decades, several researchers \cite{dvzeroski2001relational, koller2007introduction} pay lots of attention on this field. Cao et.al \cite{cao2020spectral} represent both intra-series and inter-series correlations in the spectral domain for time-series forecasting task. Li et.al \cite{li2020evolvegraph} propose a dynamic mechanism to infer the evolving latent graph for trajectory forecasting. Fan et.al \cite{fan2020self} infer the temporal relationships by sampling the time pieces from the anchor samples in the scenario of self-supervised learning. Kipf \cite{kipf2018neural} propose the neural relation inference model to infer interactions while simultaneously learning the dynamics purely from observational data under the framework of variation auto-encoder \cite{kingma2013auto}. 

In this paper, we reconstruct and align the relationships of time-series data for unsupervised domain-adaptation for time-series domain adaptation, which simultaneously considers the domain-invariant sparse associative structures and domain-variant weights.

\section{Sparse Associative Structure Alignment}\label{sec:model}

\textcolor{black}{In this paper, we first provide the problem definition of time-series domain adaptation, then we introduce the proposed \textbf{SASA-IV}.} \textcolor{black}{Our \textbf{SASA-IV} method is motivated by the process from stable causality assumption to relaxed stable association assumption. }
Under this intuition, we devise a unified model to align the invariant unweighted sparse associative structure and encode the variant strengths for time-series domain adaptation.

\subsection{Problem Formulation and Model Overview}
In this subsection, we first formulate the problem of time-series domain adaptation. Then we provide the overview of the proposed SASA-IV model.

We let $\bm{x}=\{\bm{x}_{t-N+1},\cdots, \bm{x}_{t-1},  \bm{x}_t\}$ denote a multivariate time series sample with $N$ time steps, where $\bm{x}_t\in\mathbb{R}^M$, and $y\in\mathbb{R}$ is the corresponding label. 
We assume that $P_S(\bm{x},y)$ and $P_T(\bm{x},y)$ are different distributions from the source and the target domains but are generated from a shared causal mechanism. 
Since the two variable sets generated by the same causal structure should share the same associative structure, $P_S(\bm{x},y)$ and $P_T(\bm{x},y)$ share the same associative structure. 
$(\mathcal{X}_S, \mathcal{Y}_S)$ and $(\mathcal{X}_T, \mathcal{Y}_T)$, which are sampled from $P_S(\bm{x},y)$ and $P_T(\bm{x},y)$ respectively, denote the source and target domain dataset. 
In unsupervised domain adaptation, each source domain sample $\bm{x}_S$ comes with $y_S$, while the target domain has no labeled sample. Our goal is to devise a predictive model that can predict $y_T$ given time series sample $\bm{x}_T$ from the target domain.

\begin{figure*}[t]
		\centering
		\includegraphics[width=2\columnwidth]{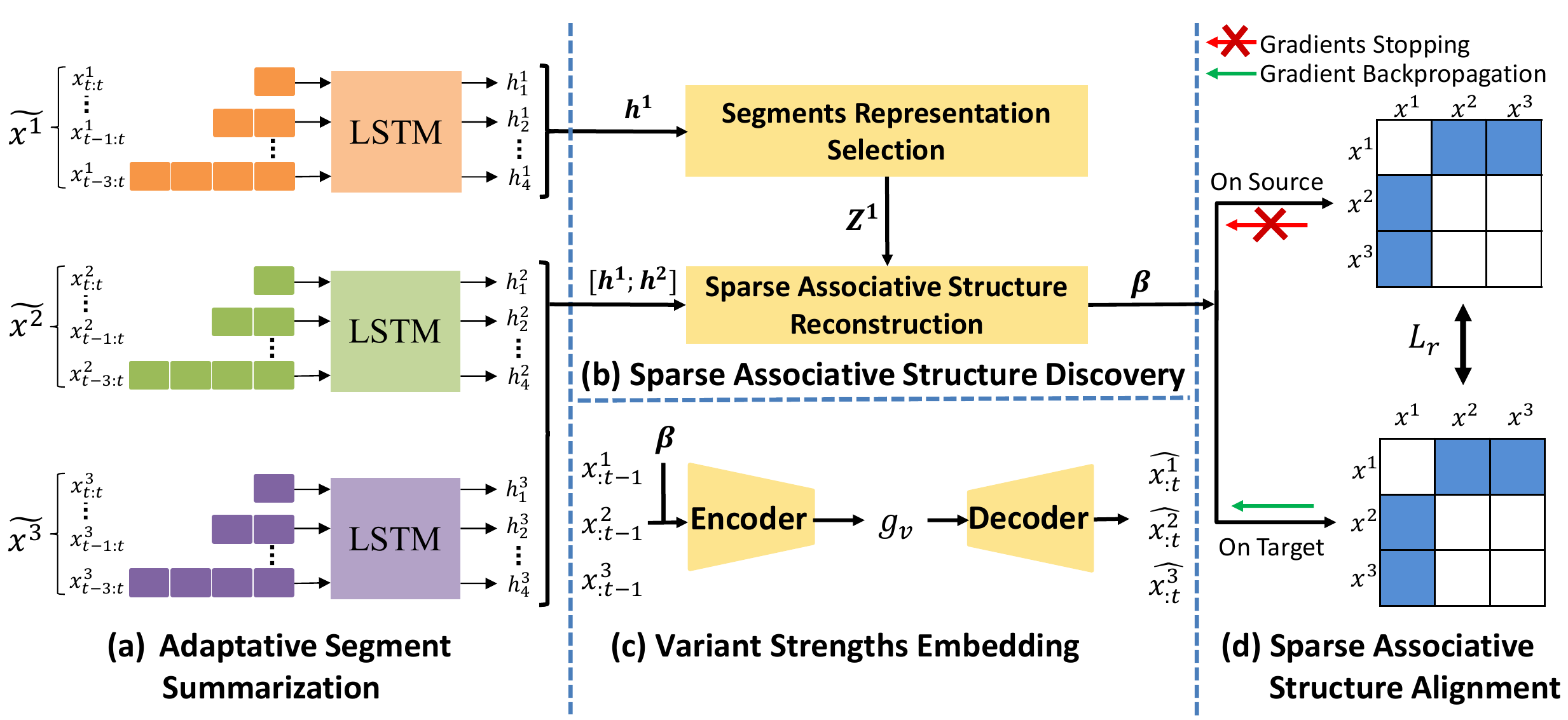}
		\caption{The framework of the SASA-IV model. (a) Adaptative segment summarization process with variable-specific LSTM. (b) Unweighted sparse associative structure discovery via intra-variables and inter-variables attention mechanism. (c) Variant Strength Embedding module to encode the domain-variant strength information. (d) Unidirectional sparse associative structure alignment between the source and the target domain. (\textit{Best view in color.})}
		\label{fig:model_1}
\end{figure*}

Based on the aforementioned problem definition, we aim to extract the domain-invariant structure information and encode the domain-variant strength information under a unified framework of sparse associative structures. The solution is inspired by the intuition that the causal mechanism is invariant across different domains. Due to the complexity of discovering causal structures, we relax the causal structures to the sparse associative structures. Considering that the offsets vary with different domains and hinder the model from extracting the domain-invariant associative structures, we first elaborate on how to obtain the fine-grain segments of time-series data to ease the obstacle of the offsets. Second, considering time lags from different domains, we reconstruct the unweighted associative structures with the help of the intra-variables and the inter-variables attention mechanisms. Different from the existing works that align the feature from different domains, the proposed method employs the unidirectional sparse associative structure alignment restriction to obtain the common associative structures from different domains to indirectly extract the domain-invariant representation. We encode the domain-variant strength information with the help of domain-variant autoencoder.


\subsection{Adaptive Segment Summarization}
In this subsection, we will elaborate on how to obtain the candidate segments to remove the obstacle of offsets. As shown in Figure \ref{fig:motivation1}, the orange blocks, whose duration varies with different domains, denote the segment of the change of variable `G'. Existing methods, which take the whole time-series data as input, can not accurately capture when a segment stars and when a variable affects the others, i.e., the sphere of influence of any variables. Therefore, these methods can not address the obstruction of offsets (i.e., the duration between the start point of time-series and the start point of a segment).

To address the this problem, we first propose the adaptive segment summarization, which is shown in Figure \ref{fig:model_1}(a). To obtain the candidate segments of $i$-th time-series $\Tilde{\bm{x}}^i$, we construct multiple segments with different length for each variable $\bm{x}^i$. Hence, we let $\Tilde{\bm{x}}^i$ be the segment set of $\bm{x}^i$ shown as Equation (1):
\begin{equation}
\small
    \Tilde{\bm{x}}^i=\{\bm{x}^i_{t:t}, \bm{x}^i_{t-1:t}, \cdots, \bm{x}^i_{t-\tau+1:t}, \cdots, \bm{x}^i_{t-N+1:t}\},
\end{equation}

Motivated by RIM \cite{goyal2019recurrent}, we allocate an independent LSTM for each variable. In detail, given a segment of $i$-th variable with $\tau$ timestamps, we have:
\begin{equation}
\small
    \bm{h}_{\tau}^i = f_i(\bm{x}^i_{t-\tau+1;t};\bm{\theta}^i),
\end{equation}
in which $f_i(\cdot)$ denote the $i$-th long term short term memory (LSTM) for $\bm{x}^i$ and $\bm{\theta}^i$ denote the parameters of $f_i(\cdot)$. For convenience, we let $\Theta=\{\theta^1, \cdots, \theta^i, \cdots, \theta^M\}$ be parameters of all the LSTM. Note that the segments in the same segment set $\Tilde{\bm{x}}^i$ share the same LSTM. And finally we can obtain the segments representation set shown as follow:
\begin{equation}
\small
    \bm{h}^i=\{\bm{h}_1^i, \cdots, \bm{h}_2^i, \cdots,\bm{h}_N^i\}.
\end{equation}

Since it is almost impossible to consider all the exact segments from the multivariable time-series data, we first obtain the representation of all candidate segments via the aforementioned processing. The most suitable segment representation are selected and used to reconstruct the associative structure, which will be described in the following subsections.

\subsection{Sparse Associative Structure Discovery}\label{sec:invariant}
In this section, we will introduce how to generalize the most exact segment representation and how to reconstruct the associative structure with the help of intra-variable attention mechanism and inter-variable attention mechanism respectively.

\subsubsection{Segments Representation Selection via Intra-Variables Attention Mechanism.}
In order to get rid of the obstacle brought from the offsets, we need to pay more attention to the exact segment representation among all the candidate segment representations with the help of the self attention mechanism \cite{vaswani2017attention}. Formally, we calculate the weights of each segment of $\bm{x}^i$ as follow:
\begin{equation}
\small
\begin{split}
\bm{\alpha}^i=&[\alpha_1^i, \cdots, \alpha_{\tau}^i,\cdots,\alpha_N^i]\\=&\text{sparsemax}\left([u_1^i, \cdots, u_{\tau}^i,\cdots,u_N^i] \right),\\
    u_{\tau}^i=&\frac{1}{N}\sum_{k=1}^N\frac{\left(\bm{W}^Q\bm{h}_{\tau}^i\right)^\mathsf{T}\left(\bm{W}^K\bm{h}_{\tau}^i\right)}{\sqrt{d_h}},
\end{split}
\end{equation}
in which $\bm{W}^Q$ and $\bm{W}^K$ are the trainable projection parameters and $\sqrt{d_h}$ is the scaling factor. In order to obtain the sparse weights that represent specific segment representation clearly, we employ sparsemax \cite{martins2016softmax} to calculate the weights. The sparsemax is defined as:
\begin{equation}
    \text{sparsemax}(\bm{z})=\mathop{\arg \min_{\bm{p} \in  {\Delta}^{K-1}}}{||\bm{p}-\bm{z}||}^2,
\end{equation}
which returns the Euclidean projection of vector $\bm{z}\in \mathbb{R}^K$ onto probability simplex ${\Delta}^{K-1}$.

\subsubsection{Sparse Associative Structure Reconstruction via Inter-variables Attention Mechanism.}
With the help of the intra-variables attention mechanism, we can extract the weighted segment representations despite the obstacle of offsets, which is shown as follow:
\begin{equation}
\small
\label{equ:feature1}
    \bm{Z}^i=\sum_{\tau=1}^N \alpha_{\tau}^i \left(\bm{W}^V \bm{h}_{\tau}^i\right), 
\end{equation}
in which $\bm{W}^V$ is trainable projection parameter. Note that $\alpha$ also denotes the probability of the length of a segment. 

Then we leverage these weighted segment representation to reconstruct the sparse associative structure among variables. So we propose the inter-variables attention mechanism to mine the associative structure among variables.

In this part, our goal is to reconstruct the associative structure among variables. Technologically, we employ the standard attention mechanism \cite{DBLP:journals/corr/BahdanauCB14}. One of the most straightforward methods to calculate the degree of correlation of variable $i$ and variable $j$ is shown as follow:
\begin{equation}
\small
\label{equ:eij}
    \bm{e}^{ij}=\frac{\bm{Z}^i\cdot \bm{Z}^j}{||\bm{Z}^i||\cdot||\bm{Z}^j||}.
\end{equation}

However, the associative structure calculated by Equation (\ref{equ:eij}) ignores the time lags from different domains between $i$-th and $j$-th variables, \textcolor{black}{which may result in the false estimation for the associative structures.} In order to take the time lags into consideration, we calculate the degrees of association between $i$-th variable and $j$-th variable by: 
\begin{equation}
\small
\begin{split}
    e_{\tau}^{ij}&=\frac{\bm{Z}^i\cdot \bm{h}_{\tau}^j}{||\bm{Z}^i||\cdot|| \bm{h}_{\tau}^j||},\\
    \bm{e}^{ij}&=\{e_1^{ij},\cdots,e_{\tau}^{ij},\cdots,e_N^{ij}\},
\end{split}
\end{equation}
Then we normalized these degrees of association with Sparsemax \cite{martins2016softmax}. Formally, we have:
\begin{equation}
\small
    \label{equ:beta}
    \begin{split}
        \bm{\beta}^i&=[\bm{\beta}^{i1}, \cdots, \bm{\beta}^{ij},\bm{\beta}^{iM}]\\&=\text{sparsemax}\left([\bm{e}^{i1},\cdots,\bm{e}^{ij}, \cdots,\bm{e}^{iM}]\right)(j\neq i).
    \end{split}
\end{equation}
Note that $\beta^{ij}_{\tau}\in\bm{\beta}^i$ denotes the associative strength between $i$-th variables and $j$-th variables with regard to segment duration of $\tau$.

Similar to Equation (\ref{equ:feature1}), we calculate the associative structure representation of the $i$-th variable as follows:
\begin{equation}
\small
    \begin{split}
        &\bm{U}^{ij} = \sum_{\tau=1}^{N}{\beta}^{ij} \cdot {\bm{h}}^j_\tau ,\\
        &{\bm{U}}^i = \sum_{m=1, m \neq i}^{M}\bm{U}^{im}.
    \end{split}
\end{equation}

\subsection{Sparse Associative Structure Alignment}

\subsubsection{Unweighted Sparse Associative Structure Alignment.} 

Based on the associative structures that are extracted in Equation (\ref{equ:beta}), we need to restrict the distance of the structure between the source and the target domains for extracting the domain-invariant associative structures. The conference version borrows the idea of domain confuse network and restricts the distance of $\bm{\beta}$ with the help of maximum mean discrepancy (MMD) \cite{long2015learning}, which is shown in Equation \ref{equ:mmd_resct_beta}.
\begin{equation}
\small
    \mathcal{L}_{\beta} = \textbf{MMD}(\bm{\beta}_S,\bm{\beta}_T).
    \label{equ:mmd_resct_beta}
\end{equation}

Since the time lags between any two variables may be similar but different, the duration of segments might change over different domains. Therefore, in order to reconstruct the associative structure more precisely, we minimize the \textbf{MMD} between $\bm{\alpha}$ from the source and the target domain to align the offsets. It restricts the duration of the segment from different domains to be similar, which contributes to extracting structure for transfer. 
Formally, we have:
\begin{equation}
\small
    \begin{split}
        \mathcal{L_{\alpha}}=\text{MMD}(\bm{\alpha}_S,\bm{\alpha}_T).
    \end{split}
    \label{equ:mmd_resct_alpha}
\end{equation}

Though the associative structures are stable, there are some domain-variant factors like the weights of the associative structures. Furthermore, domain-variant weights play an important role in forecasting. For example, we let gravity and mass be composed of an associative structure, the value of gravity also depends on the gravitational acceleration, which denotes the weights of the associative structure. 
Straightforwardly minimizing the distance between the structures with strengths will sacrifice the domain-variant information, which further degenerates the model performance.
In order to address the aforementioned problem, we separate the weights from the associative structures and minimize the discrepancy between the unweighted structures from different domains, hence we modify Equation (\ref{equ:mmd_resct_beta}) and (\ref{equ:mmd_resct_alpha}) to the Equation (\ref{equ:alpha}):
\begin{equation}
\small
\label{equ:alpha}
    \begin{split}
    \mathcal{L_{\beta}}&=||\mathbbm{1}(\bm{\beta}_S>\mu)-\mathbbm{1}(\bm{\beta}_T>\mu)||_1,\\
        \mathcal{L_{\alpha}}&=||\mathbbm{1}(\bm{\alpha}_S>\mu)-\mathbbm{1}(\bm{\alpha}_T>\mu)||_1,
    \end{split}
\end{equation}
in which $||\cdot||_1$ denotes the L1 norm and the indicator function $\mathbbm{1}(\cdot)$ is used to choose the edges with weights greater than $\mu$. We also find that changing the value $\mu$ can control the sparseness of the generated associative structures, so we can remove more redundant relationships by tuning $\mu$.

\subsubsection{Unidirectional alignment restriction.} As mentioned in Equation (\ref{equ:alpha}), we extract the unweighted associative structures with the help of attention mechanisms, which are essential profited from the labeled source data. And the target associative structures might be wrongly extracted without any supervised signal. If we directly employ Equation (\ref{equ:alpha}) to align the structures, the wrong structures from the target domain might be aligned to the graph of source domain. In the worst case, the source structures would be totally wrong, which might result in the negative transfer.

In order to address the this issue, we provide a simple but effective solution that prevent the knowledge transfer from target to source. In detail, we apply the gradient stopping operation $\mathcal{C}(\cdot)$ on the source associative structures, which is shown as follows:
\begin{equation}
\small
\label{equ:alpha2}
    \begin{split}
    \mathcal{L_{\beta}}&=||\mathcal{C}(\mathbbm{1}(\bm{\beta}_S>\mu))-\mathbbm{1}(\bm{\beta}_T>\mu)||_1,\\
        \mathcal{L_{\alpha}}&=||\mathcal{C}(\mathbbm{1}(\bm{\alpha}_S>\mu))-\mathbbm{1}(\bm{\alpha}_T>\mu)||_1,
    \end{split}
\end{equation}

Note that L1 Norm of a matrix is calculated by $||A||_1=\sum_{ij}|A_{ij}|$, where $A_{ij}$ is the $i$-th row $j$-th column element in matrix A.

\subsection{Domain-variant information Extraction}
In subsection \ref{sec:invariant}, we have discussed that the domain-invariant unweighted associative structures are domain-invariant while some domain-variant factors play an important role in model prediction.

However, it is not a simple task to extract the domain-variant information from the source to the target domain, one major difficulty comes from the unlabeled target domain data, since we can only extract the source-specific information with the help of source labeled data. 
Fortunately, the stationary time-series data, which are generated via a stable causal structure, contain the inherent autoregressive property. This inspiration enlightens us to extract the domain-variant factors in a autoregressive paradigm. Hence, we devise the domain-variant factor extraction module, which is a graph attention networks \cite{velivckovic2017graph,wang2019heterogeneous,kipf2016semi} based autoregressive autoencoder. Formally, \textcolor{black}{the encoder and the decoder are respectively represented as}:
\begin{equation}
\small
\begin{split}
    \bm{g}_v = \textbf{GNN}(&\mathbbm{1}(\bm{\beta}>\mu)), \psi_e(\bm{x}_{1:t-1});\bm{W}^e),
\end{split}
\label{equ:encoder}
\end{equation}
\begin{equation}
\begin{split}
    \hat{\bm{x}_t}=&\psi_d(\bm{g}_v;\bm{W}^d),\\
\end{split}
\label{equ:decoder}
\end{equation}
in which $\bm{W}^g$ and $\bm{W}^d$ are the trainable parameters.

In the encoder, we first use the LSTM-based feature extractor $\psi_e(\cdot)$ to extract the feature for each variable. Note that we allocate an independent LSTM for each variable. Then we take the unweighted sparse associative structures and the $\bm{x}_{1:t-1}$ and the extracted feature as the input of graph attention networks and obtain domain-variant representation $\bm{g}_v$. In the decoder, we use $\bm{g}_v$ to generate the predicted result $\hat{\bm{x}_t}$.
$\hat{\bm{x}_t}$ are the predicted future values. Finally, we use mean squared error (MSE) to optimize the autoregressive model, which is shown as follows:
\begin{equation}
\small
    \mathcal{L}_r=MSE(\hat{\bm{x}_t}, \bm{x}_t).
    \label{equ:l_r}
\end{equation}

In summary, the GNN work as the encoder which approximately imitate the the data generation process under the invariant sparse associative structures; $\psi_d(\cdot)$ works as the decoder and predict the value in the final timestamp to guarantee the the domain-variant factors are preserved. So we can extract the source and target domain-variant factors by respectively using the source and target training data.


\subsection{Model Summary}
\subsubsection{Task based Label Predictor.}
Finally, we can obtain the final representation that contains the domain-invariant associative structure information and the domain-variant strength information as shown in Equation (\ref{equ:con}):
\begin{equation}
\small
\label{equ:con}
\begin{split}
    &\bm{H}^i=\bm{Z}^i\oplus\bm{U}^i,\\
    &\hat{\bm{H}}=\bm{H}^1 \oplus \bm{H}^2 \oplus \cdots \oplus \bm{H}^M,\\
    &\bm{H} = \hat{\bm{H}} \oplus \bm{g}_v,
\end{split}
\end{equation}
\textcolor{black}{in which $\oplus$ denotes the concatenate operation.}

For convenience, we describe the above process as :
\begin{equation}
\small
    \bm{H}=G_f(\bm{x};\Theta, \bm{W}^Q,\bm{W}^K, \bm{W}^V, \bm{W}^e, \bm{W}^d).
\end{equation}

After obtaining the final representation, we take $\bm{H}$ as the input of label classifier $G_y(\cdot;\bm{\phi})$ whose loss function is $\mathcal{L}_y$. For the classification problems, we employ cross-entropy as the label loss. For the regression problems, we employ RMSE as the label loss.


\subsubsection{Objective Function.} The total loss of the proposed structure alignment model for time series domain adaptation is formulated as:
\begin{equation}
\small
	\begin{split}
	\mathcal{L}\left(\Theta, \bm{W}^Q, \bm{W}^K,\bm{W}^V, \bm{W}^e, \bm{W}^d, \bm{\phi}\right) = \\\mathcal{L}_y + \omega (\mathcal{L}_{\alpha} + \mathcal{L}_\beta) + \gamma \mathcal{L}_r,
	\end{split}
\end{equation}
in which $\omega$ and $\gamma$ are hyper-parameters. 

Under the above objective function, our model is trained on the source and target domain using the following procedure:
\begin{equation}
    \small
	\begin{split}
	\mathop{\arg \min_{\bm{\Theta}, \bm{W}^Q, \bm{W}^K, \bm{W}^V, \bm{W}^e, \bm{W}^d, \bm{\phi}}}\mathcal{L}\left(\bm{\Theta}, \bm{W}^Q, \bm{W}^K,\bm{W}^e, \bm{W}^d, \bm{W}^V, \bm{\phi}\right).
	\end{split}
\end{equation}

\section{Theoretical Analysis}\label{sec:theore}

Several works have focused on the generalization theory of domain adaptation \cite{mohri2018foundations,zhang2019bridging,mansour2009domain,cortes2011domain,ben2007analysis,mansour2009domain}, which mainly depends on the distance of distribution between the source and the target domains. In this paper, we first provide the straightforward definition of time-series structural distance based on the generation process of time-series data, then establish the generalization bound for time-series unsupervised domain adaptation, in which the generalization risk on the target domain depends on not only the risk on the source data but also the time-series structural distance between the source and the target domains. 

\subsection{Structural Generation Distance for Time-Series Data.}
In this subsection, we provide the heuristic definition of Structural Distance (\textbf{SD}) for time-series data. We first introduce the data generation process for time-series data based on structural causal models (\textbf{SCM}), which is shown as follows:
\begin{equation}
\small
    \label{equ:scm}
    x_t^i = F_i(\bm{pa}(x_t^i), N_i),
\end{equation}
in which $F_i$ denotes any flexible types of function for $i$-th time-series data $x^i$; $\bm{pa}(x_t^i)$ denotes the parents of $x_t^i$; and $N_i$ denotes the independent noise terms. According to Equation (\ref{equ:scm}), we can easily find that the distribution of $x_t^i$ depends on the distribution of $\bm{pa}(x_t^i)$ and the causal mechanism.

Inspired by the aforementioned structural causal models, it is straightforward to define the distance for time-series data. As an effective metric for time-series data, the new distance should satisfy all necessary axioms for a general metric.
\begin{definition}
(\textbf{Structural Distance, SD}.) We let $\bm{x}$ be the multivariate time series, and $\bm{x}_0$ be the values of the first timestamps. We further assume that $\bm{x}$ is generated by causal structures $A$ with the strength $\mathcal{\bm{W}}$. Given $\bm{x}_S$ and $\bm{x}_T$ from the source and the target domain, the structural distance between $\bm{x}_S$ and $\bm{x}_T$ can be formalized as follows:
\begin{equation}
\small
\begin{split}
\label{equ:sgd_dis}
\textup{dis}_{\text{SD}}^{\mathcal{S}\leftrightarrow\mathcal{T}}(\bm{x}_S,\bm{x}_T)&=dist(P_S(\bm{x}_0),P_T(\bm{x}_0)) + ||A_S-A_T||_1 \\&+ ||\mathcal{\bm{W}}_S-\mathcal{\bm{W}}_T||_1.
\end{split}
\end{equation}
\end{definition}
Note that $dist(\cdot,\cdot)$ in Equation (\ref{equ:sgd_dis}) can be any distance metric for distribution, and we employ the total variation distance in the following generalization bound; $A_S$ and $A_T$ can be the sparse associative structures corresponding to their causal structures.

According to Equation (\ref{equ:sgd_dis}), we can find that the distance of time-series distribution is affected by the following three factors:
\begin{itemize}
    \item The first term $dist(P_S(\bm{x}_0),P_T(\bm{x}_0))$ denotes the distribution distance of the start points $\bm{x}_0$. An comprehensible example is that the value ranges of time-series data influence the distribution. Note that $dist(\cdot,\cdot)$ denotes any type of distribution metric for static data.
    \item The second term $||A_S-A_T||_1$ denotes the distance between the unweighted sparse associative structures from the source and target domains.
    \item The third term $||\mathcal{\bm{W}}_S-\mathcal{\bm{W}}_T||_1$ denotes the distance between the strengths of causal structures from the source and target domains.
\end{itemize}

Moreover, we find that the structural generation distance satisfies the three axioms for a general metric:
\begin{theorem}
\label{theor:dist}
The Structural Distance (\textbf{SD}) satisfies the three axioms for a general metric, to be specific, it satisfies the following conditions:\\

(1) $\textup{dis}_{\text{SD}}^{\mathcal{S}\leftrightarrow\mathcal{T}}\geq0$ and $\textup{dis}_{\text{SD}}^{\mathcal{S}\leftrightarrow\mathcal{T}}=0$ if and only if $\mathcal{S}=\mathcal{T}$;\\

(2) $\textup{dis}_{\text{SD}}^{\mathcal{S}\leftrightarrow\mathcal{T}}(\bm{x}^S, \bm{x}^T)=\textup{dis}_{\text{SD}}^{\mathcal{S}\leftrightarrow\mathcal{T}}(\bm{x}^T,\bm{x}^S)$ (symmetric);\\

(3) $\textup{dis}_{\text{SD}}^{\mathcal{S}\leftrightarrow\mathcal{T}}\leq \textup{dis}_{\text{SD}}^{\mathcal{S}\leftrightarrow\mathcal{D}}+\textup{dis}_{\text{SD}}^{\mathcal{D}\leftrightarrow\mathcal{T}}$ (triangle inequality).
\end{theorem}

Theorem \ref{theor:dist} are easy to be proved, it is not hard to find that time-series structurally generation distance can be used to measure the distance between the distribution of time-series data.

\subsection{Generalization Bounds}

In this section, we provide the generalization bound for time-series unsupervised domain adaptation. We first formalize some notations that will be used in the following statement. Suppose $\mathcal{X}$ be an instance set of time-series and $\{0,1\}$ be the label set for binary classification. We let $\mathcal{H}$ be a hypothesis space that maps $\bm{x}$ to $\mathbb{R}$ and $\forall h \in \mathcal{H}, h:\mathcal{X}\rightarrow \{0,1\}$. We further let $\eta: \mathcal{X}\rightarrow \{0,1\}$ be the labeling function. The probability according to the distribution $P_\mathcal{S}$ is defined as $\epsilon_\mathcal{S}(h,f)=\mathbb{E}_{\bm{x}\sim P_\mathcal{S}}[|h(\bm{x})-\eta_S(\bm{x})|]$. We use the shorthand $\epsilon_S(h)=\epsilon(h,\eta_S)$ and $\epsilon_T(h)$ is defined the same. Based on the aforementioned definition, we make the following assumption:
\begin{assumption}
\label{ass1}
\textbf{Time-series distribution bound assumption:}
Suppose that $P_S$ and $P_T$ are the source and the target distributions and $A_S$ and $A_T$ are sparse associative structures with strengths $\mathcal{W}_S, \mathcal{W}_T$, then a positive value $K$ exists that makes the following inequality hold:

\begin{equation}
\begin{split}
\small
    |P_\mathcal{S}(\bm{x})-P_\mathcal{T}(\bm{x})| \leq& K (|P_\mathcal{S}(\bm{x}_0)-P_\mathcal{T}(\bm{x}_0)|+||A_S-A_T||_1 \\&+ ||\mathcal{\bm{W}}_S-\mathcal{\bm{W}}_T||_1)
    \\=& K\textup{dis}_{\text{SD}}^{\mathcal{S}\leftrightarrow\mathcal{T}}(\bm{x}_S,\bm{x}_T),
\end{split}
\end{equation}
\end{assumption}

Note that the third line establishes when $dist(P_S(\bm{x}_0),P_T(\bm{x}_0))$ is the total variation distance.

Based on the aforementioned definition and assumption, we propose the generalization bound of the propose \textbf{SASA-IV}, which is shown as follows.

\begin{theorem} (\textbf{Generalization Bound for Time-Series Unsupervised Domain Adaptation.})
\label{bound}
Given Assumption \ref{ass1}, we have:
\begin{equation}
\small
\label{equ:bound}
\begin{split}
    \epsilon_T(h) \leq & \epsilon_S(h) + K(|P_\mathcal{S}(\bm{x}_0)-P_\mathcal{T}(\bm{x}_0)|+||A_S-A_T||_1 \\&+ ||\mathcal{\bm{W}}_S-\mathcal{\bm{W}}_T||_1) + \lambda,\\
    = &\epsilon_S(h) + K\textup{dis}_{\text{SD}}^{\mathcal{S}\leftrightarrow\mathcal{T}}(\bm{x}_S,\bm{x}_T) + \lambda
\end{split}
\end{equation}
in which $K, \lambda$ are constants and $h \in \mathcal{H}$ a any hypothesis.
\begin{proof}
\begin{equation}
\small
\begin{split}
    &\epsilon_T(h)=\epsilon_T(h) + \epsilon_S(h) - \epsilon_S(h) + \epsilon_S(h, \eta_T) - \epsilon_S(h, \eta_T) \\
    \leq& \epsilon_S(h) + \epsilon_S(h, \eta_T)-\epsilon_S(h, \eta_S) + |\epsilon_T(h)-\epsilon_S(h, \eta_T)|\\
    \leq& \epsilon_S(h) + \epsilon_S(\eta_S, \eta_T) + |\epsilon_T(h)-\epsilon_S(h, \eta_T)|\\
    \leq& \epsilon_S(h) + \epsilon_S(\eta_S, \eta_T) + \int \!\! |P_S(\bm{x})\!-\!P_T(\bm{x})||h(\bm{x})-\eta_T(\bm{x})|d\bm{x}\\
    \leq& \epsilon_S(h) + \epsilon_S(\eta_S, \eta_T) + \int \!\! |P_S(\bm{x})\!-\!P_T(\bm{x})|d\bm{x}\\
    \leq& \epsilon_S(h) + K(|P_\mathcal{S}(\bm{x}_0)-P_\mathcal{T}(\bm{x}_0)|+||A_S-A_T||_1 \\&+ ||\mathcal{\bm{W}}_S-\mathcal{\bm{W}}_T||_1) + \epsilon(\eta_S, \eta_T)\\
    =& \epsilon_S(h) + K\textup{dis}_{\text{SD}}^{\mathcal{S}\leftrightarrow\mathcal{T}}(\bm{x}_S,\bm{x}_T) + \lambda
\end{split}
\end{equation}
in which $\lambda= \epsilon(\eta_S, \eta_T)$ is a constant.
\end{proof}

\end{theorem}
According to this generalization bound, we can find that the expected risk $\epsilon_T(h)$ is not only controlled by $\epsilon_S(h)$ but also the distance between the unweighted associative structures, the distance between the strengths as well as the structural generation distance between the source and the target domains. It inspires us to extract the domain-invariant unweighted structures and weights, which is essentially employed in the conference version.
In the meanwhile, the proposed \textbf{SASA-IV} takes the domain-variant factors into consideration, so the strengths alignment term $||\mathcal{\bm{W}}_S-\mathcal{\bm{W}}_T||_1$ and the start point distribution alignment term $|P_\mathcal{S}(\bm{x}_0)-P_\mathcal{T}(\bm{x}_0)|$ can be removed. Hence we can obtain a tighter bound, which is derived Equation (\ref{equ:bound}) as follows:
\begin{equation}
    \epsilon_T(h) \leq \epsilon_S(h) + K||A_S-A_T||_1 + \lambda.
\end{equation}

\section{Experiments}\label{sec:exp}
\subsection{Dataset}
\subsubsection{Air Quality Forecast Dataset.}
The air quality forecast dataset\cite{zheng2015forecasting} is collected in the Urban Air project\footnote{https://www.microsoft.com/en-us/research/project/urban-air/} from 2014/05/01 to 2015/04/30, which contains air quality data, meteorological data, weather forecast data, etc. The dataset covers 4 major Chinese cities: Beijing (B), Tianjin (T), Guangzhou(G), and Shenzhen(S). We employ air quality data as well as meteorological data to predict PM2.5. We choose the air quality station with the least missing value and take each city as a domain. We use this dataset because the air quality data is common and the sensors in the smart city systems usually contain complex causality. The associations among sensors are often sparse, which is suitable for our model. 

	
	
\subsubsection{In-hospital Mortality Prediction Dataset.}
MIMIC-III\cite{johnson2016mimic,che2018recurrent}\footnote{https://mimic.physionet.org/gettingstarted/demo/} is another published dataset with de-identified health-related data associated with more than forty thousand patients who stayed in critical care units of the Beth Israel Deaconess Medical Center between 2001 and 2012. It's the benchmark of time series domain adaptation in VRADA\cite{DBLP:conf/iclr/PurushothamCNL17}.
Similar to Purushotham et al.\cite{purushotham2018benchmarking}, we choose 12 time series (such as Heart Rate, Temperature, Systolic blood pressure, etc) and five static feature from 35637 records. In order to prepare the in-hospital mortality prediction dataset for time series domain adaptation, we split the patients into 4 groups according to their age (Group1: 20-45, Group2: 46-65, Group3: 66-85, Group4: \textgreater 85).

	

\subsubsection{Boiler Fault Detection Dataset.}
The boiler data consists of sensor data from three boilers from 2014/3/24 to 2016/11/30. There are 3 boilers in this dataset and each boiler is considered as one domain. The learning task is to predict the \textit{faulty blowdown valve} of each boiler. Since the fault data is very rare. It's hard to obtain the fault samples in the mechanical system. So it's important to utilize the labeled source data and unlabeled target data to improve the model generalization.
	

\subsection{Compared Methods}
We consider as many as possible compared methods for time-series domain adaptation, which are described as follows:
\begin{itemize}
    \item \textbf{LSTM\_S2T.} LSTM\_S2T uses the source domain data to train a vanilla LSTM model and applies it to the target domain without any adaptation (S2T stands for source to target). It's expected to provide the lower bound performance.
    \item \textbf{R-DANN.} R-DANN \cite{da2020remaining} is an unsupervised domain adaptation architecture proposed in \cite{ganin2015unsupervised} with GRL (Gradient Reversal Layer) on LSTM, which is a straightforward solution for time series domain adaptation.
    \item \textbf{RDC.} Deep domain confusion is an unsupervised domain adaptation method proposed in \cite{tzeng2014deep} which minimizes the distance between the source and target distributions by employing Maximum Mean Discrepancy (MMD). Similar to the aforementioned R-DANN, we use LSTM as the feature extractor for time series data.
    \item \textbf{VRADA.} VRADA \cite{DBLP:conf/iclr/PurushothamCNL17} is a time series unsupervised domain adaptation method which combines the GRL and VRNN \cite{chung2015recurrent}.
    \item \textbf{AdvSKM} \cite{ijcai2021-378} is one of the latest approaches for time-series domain adaptation. The \textbf{AdvSKM} uses the hybrid spectral kernel network to reform the MMD metric.
    \item \textbf{CODATS} \cite{wilson2020multi} leverages the convolutional neural networks and weak supervision in the form of target-domain label distribution. 
\end{itemize}

\subsection{Model Variants}
In order to verify the effectiveness of each component of our model, we further devise the following model variants.
	\begin{itemize}
	    \item SASA: We take the previous SASA in conference as a model variant.
	    \item SASA-$\bm{\alpha}$: We remove $\mathcal{L}_\alpha$ to verify the usefulness of the segment length restriction loss from the SASA model.
	    \item SASA-$\bm{\beta}$: We remove $\mathcal{L}_\beta$ to verify the usefulness of the sparse associative structure alignment loss from the SASA model.
    	\item SASA-IV-$\bm{\alpha}$: We remove $\mathcal{L}_\alpha$ to verify the usefulness of the segment length restriction loss.
	    \item SASA-IV-$\bm{\beta}$: We remove $\mathcal{L}_\beta$ to verify the usefulness of the sparse associative structure alignment loss.
	    \item SASA-IV-$\bm{\gamma}$: We remove the $\mathcal{L}_\gamma$ to verify the usefulness of the domain-variant factors.
	    \item SASA-IV-$\bm{C}$: We remove the gradient stopping operation $\mathcal{C}(\cdot)$ to verify the usefulness of the unidirectional alignment restriction.
	    
\end{itemize}
\begin{table*}
	\small
	\caption{RMSE on air quality prediction.}
		\centering
		\resizebox{\textwidth}{22mm}{
		    \scalebox{0.95}{
			\begin{tabular}{l|lllllllllllll}
				\toprule[1pt]
				Method&B$\rightarrow$T&G$\rightarrow$T&S$\rightarrow$T&T$\rightarrow$B&G$\rightarrow$B&S$\rightarrow$B&B$\rightarrow$G&T$\rightarrow$G&S$\rightarrow$G&B$\rightarrow$S&T$\rightarrow$S&G$\rightarrow$S&Avg \\
				\toprule[1pt]
				LSTM\_S2T  	           & 40.89 & 42.20 & 49.21 & 52.18 & 56.55 & 70.52 & 19.12 & 19.37 & 17.53 & 13.82 & 13.80 & 14.17 & 34.11 \\ 
				RDC            		   & 39.31 & 40.36 & 47.75 & 51.99 & 56.77 & 69.40 & 18.84 & 19.28 & 15.56 & 13.67 & 13.56 & 13.91 & 33.37 \\
				R-DANN                 & 41.49 & 39.85 & 46.76 & 52.69 & 54.80 & 68.92 & 18.11 & 18.95 & 15.14 & 13.76 & 13.86 & 13.83 & 33.18 \\
				VRADA                  & 38.56 & 39.12 & 46.12 & 52.74 & 54.57 & 65.53 & 17.84 & 18.55 & 14.75 & 13.84 & 14.22 & 13.85 & 32.50 \\
				AdvSKM                  & 40.21 & 39.23 & 46.74 & 47.14 & 55.79 & 62.74 & 18.63 & 19.45 & 17.26 & 14.06 & 16.95 & 13.67 & 32.65 \\
				CODAT                  & 38.47 & 38.70 & 48.19 & 48.17 & 55.17 & 57.65 & 17.86 & 18.38 & 17.61 & 14.76 & 17.58 & 14.16 & 32.29 \\
				\hline
				SASA             & 35.52 & 34.44 & 40.74 & 49.41 & 53.98 & 57.42 & 16.45 & 15.84 & 14.27 & 13.52 & 13.47 & \textbf{13.48} & 29.88\\
				SASA-IV             & \textbf{35.14} & \textbf{33.76} & \textbf{40.51} & \textbf{46.28} & \textbf{53.06} & \textbf{55.88} & \textbf{15.26} & \textbf{15.01} & \textbf{14.01} & \textbf{13.00} & \textbf{13.01} & 13.50 & \textbf{29.03} \\
				\toprule[1pt]
		\end{tabular}}}

		\label{tab:air_quality}
	\end{table*}
	
	\begin{table*}
		\centering
		\small
		\caption{AUC score(\%) on in-hospital mortality prediction.}
		\resizebox{\textwidth}{22mm}{
			\begin{tabular}{l|lllllllllllll}
				\toprule[1pt]
				Method&2$\rightarrow$1&3$\rightarrow$1&4$\rightarrow$1&1$\rightarrow$2&3$\rightarrow$2&4$\rightarrow$2&1$\rightarrow$3&2$\rightarrow$3&4$\rightarrow$3&1$\rightarrow$4&2$\rightarrow$4&3$\rightarrow$4&Avg \\
				\toprule[1pt]
				LSTM\_S2T 	           & 80.11 & 78.09 & 76.91 & 80.22 & 81.26 & 76.09 & 75.73 & 79.21 & 75.07 & 65.35 & 60.07 & 69.15 & 75.52 \\ 
				RDC           	   & 80.96 & 78.32 & 77.18 & 80.28 & 82.63 & 77.36 & 76.04 & 79.90 & 75.52 & 65.73 & 69.16 & 70.75 & 76.15 \\
				R-DANN         	   & 80.88 & 79.57 & 77.35 & 80.41 & 82.14 & 78.24 & 75.93 & 79.01 & 75.80 & 66.55 & 69.52 & 69.49 & 76.24 \\
				VRADA                  & 80.94 & 80.81 & 77.08 & 81.52 & 83.09 & 78.25 & 75.57 & 79.24 & 75.67 & 68.14 & 69.23 & 69.94 & 76.62 \\
				AdvSKM                  & 83.03 & 81.90 & 78.08 & 80.52 & 83.42 & 77.08 & 75.61 & 79.38 & 76.40 & 65.51 & 70.03 & 70.93 & 76.83 \\
				CODAT                   & 81.11 & 78.09 & 76.46 & 80.24 & 83.00 & 77.23 & 75.98 & 79.09 & 75.95 & 66.14 & 69.81 & 72.31 & 76.28 \\
				\toprule[1pt]
				SASA           & 84.21 & 82.68 & 80.20 & 83.14 & 84.22 & 81.87 & 77.41 & 80.56 & 78.58 & 70.84 & 72.04 & 73.20 & 79.08 \\
				
				SASA-IV                & \textbf{85.80}     & \textbf{86.25}     & \textbf{81.32} & \textbf{83.48} & \textbf{84.74} & \textbf{82.42} & \textbf{78.67} & \textbf{80.43} & \textbf{79.04} & \textbf{74.86} & \textbf{78.58} & \textbf{79.03} & \textbf{81.22} \\
				\toprule[1pt]
		\end{tabular}}
		
		\label{tab:mimic3}
	\end{table*}

\begin{table}[bhpt]

		\caption{AUC score(\%) on boiler fault detection.}
		\centering
		\begin{tabular}{p{1.4cm}|p{0.55cm}p{0.55cm}p{0.55cm}p{0.55cm}p{0.55cm}p{0.55cm}p{0.55cm}p{0.55cm}}
			\toprule[1pt]
			\small{Method}&\small{1$\rightarrow$2}&\small{1$\rightarrow$3}&\small{3$\rightarrow$1}&\small{3$\rightarrow$2}&\small{2$\rightarrow$1}&\small{2$\rightarrow$3}&\small{Avg} \\
			\toprule[1pt]
			\small{LSTM\_S2T} 	           & \small{67.04} & \small{94.50} & \small{93.23} & \small{56.06} & \small{84.71} & \small{91.14} & \small{81.17} \\ 
			\small{RDC}              	   & \small{67.17} & \small{94.65} & \small{93.17} & \small{57.30} & \small{85.59} & \small{92.45} & \small{81.71} \\
			\small{R-DANN}            	   & \small{67.26} & \small{94.88} & \small{93.57} & \small{58.14} & \small{85.54} & \small{92.41} & \small{81.97} \\
			\small{VRADA}                  & \small{67.38} & \small{94.69} & \small{93.58} & \small{58.89} & \small{84.78} & \small{92.52} & \small{81.97} \\
			\small{AdvSKM}                  & \small{68.68} & \small{94.99} & \small{93.25} & \small{59.35} & \small{86.51} & \small{91.14} & \small{82.32} \\
			\small{CODAT}                  & \small{68.62} & \small{94.75} & \small{93.31} & \small{58.38} & \small{86.03} & \small{91.14} & \small{82.04} \\
			\toprule[1pt]
			\small{SASA }                 & \small{71.56} & \small{95.39} & \small{93.33} & \small{61.90} & \small{88.04} & \small{93.16} & \small{83.89} \\
			\small{SASA-IV}                 & \small{\textbf{73.61}} & \small{\textbf{95.64}} & \small{\textbf{93.77}} & \small{\textbf{65.28}} & \small{\textbf{92.37}} & \small{\textbf{94.75}} & \small{\textbf{85.90}}\\
			\toprule[1pt]
		\end{tabular}
		
		\label{tab:boiler}
\end{table}




\subsection{Implementation Details}
In our experiments, we follow the standard protocol of unsupervised domain adaptation and leverage the labeled source data and the unlabeled target data. All the experiments are conducted on NVIDIA GeForce RTX 2070S GPU. We use a batch size of 1024 for all the datasets. We further let $\epsilon=0.08$ for all the datasets. Note that the presented experiment results are averaged over several replicated with different random seeds, so the values might be slightly different from the conference version that reports the best results.


\subsection{Result}
\subsubsection{Results on Air Quality Forecast.}
Then we further evaluate the transferability of our model on the time-series forecasting task. We consider the air quality forecast dataset since the meteorologic sensors in the smart cities system usually contain the stable causal mechanism, meaning that the associative structures are stable. So adaptively forecasting the air quality can benefit the environmental prediction. Experiment results on the air quality forecast dataset are shown in Table \ref{tab:air_quality}. 

Similar to the results in the boiler fault detection dataset, the proposed SASA-IV also achieves the best performance and outperforms all the comparison methods. According to the results, we can find that:
\begin{itemize}
    \item Compared SASA-IV with the previous SASA, we can find that almost all the results of SASA-IV are much better than that of other methods. Note that the tasks of smaller geographical distance like $S \rightarrow B$ and $B \rightarrow S$ achieve more improvement, this is because the city pairs with closer geographical distance may not only share more common associative structures but also contain similar strengths.
    \item As for the city pairs with further geographical distance like Shenzhen and Tianjin, the improvement is a bit smaller, but the performance of SASA-IV on this task is still better than that of SASA, reflecting that the domain-invariant information can benefit the performance of time-series domain adaptation.
    \item However, the improvement is not so notable when we take Guangzhou and Shenzhen as the target domain, this is because the label value ranges of these cities are much lower than the others. And the CNN-based method like CODAT and AdvSKM do not achieve ideal performance, this is because the size of the air quality forecast dataset is small.
\end{itemize}

\begin{table*}
	\small
	\caption{RMSE on air quality prediction for ablation study.}
		\centering
		\resizebox{\textwidth}{20mm}{
		    \scalebox{0.95}{
			\begin{tabular}{l|lllllllllllll}
				\toprule[1pt]
				Method&B$\rightarrow$T&G$\rightarrow$T&S$\rightarrow$T&T$\rightarrow$B&G$\rightarrow$B&S$\rightarrow$B&B$\rightarrow$G&T$\rightarrow$G&S$\rightarrow$G&B$\rightarrow$S&T$\rightarrow$S&G$\rightarrow$S&Avg \\
				\toprule[1pt]
				SASA-$\alpha$             & 36.61 & 34.71 & 41.73 & 49.92 & 54.76 & 58.53 & 17.38 & 16.36 & 14.70 & 13.87 & 13.90 & \text{13.91} & 30.53\\
				SASA-$\beta$             & 36.51 & 34.96 & 41.32 & 49.93 & 55.10 & 58.33 & 17.08 & 16.40 & 14.67 & 13.89 & 13.90 & \text{13.88} & 30.50\\
				SASA-IV-$\alpha$             & 37.60 & 34.84 & 41.67 & 48.06 & 56.83 & 59.21 & 16.30 & 15.92 & 14.37 & 14.09 & 14.11 & \text{13.99} & 30.58\\
				SASA-IV-$\beta$             & 36.47 & 35.85 & 42.40 & 47.26 & 54.42 & 59.56 & 16.78 & 16.60 & 14.63 & 13.88 & 13.85 & \text{13.80} & 30.46\\
				SASA-IV-$\gamma$             & 38.58 & 37.21 & 41.82 & 47.77 & 55.23 & 57.54 & 16.60 & 16.11 & 14.50 & 14.19 & 14.21 & \text{13.67} & 30.63\\
				SASA-IV-$C$             & 36.53 & 34.88 & 41.24 & 47.18 & 55.24 & 57.87 & 15.99 & 15.85 & 14.54 & 13.83 & 13.87 & \text{13.87} & 30.07\\
				SASA-IV             & \textbf{35.14} & \textbf{33.76} & \textbf{40.51} & \textbf{46.28} & \textbf{53.06} & \textbf{55.88} & \textbf{15.26} & \textbf{15.01} & \textbf{14.01} & \textbf{13.00} & \textbf{13.01} & \textbf{13.50} & \textbf{29.03} \\
				\toprule[1pt]
		\end{tabular}}}

		\label{tab:air_quality_ablation}
	\end{table*}

\begin{table*}
		\centering
		\small
		\caption{AUC score(\%) on in-hospital mortality prediction for ablation study.}
		\resizebox{\textwidth}{20mm}{
			\begin{tabular}{l|lllllllllllll}
				\toprule[1pt]
				Method&2$\rightarrow$1&3$\rightarrow$1&4$\rightarrow$1&1$\rightarrow$2&3$\rightarrow$2&4$\rightarrow$2&1$\rightarrow$3&2$\rightarrow$3&4$\rightarrow$3&1$\rightarrow$4&2$\rightarrow$4&3$\rightarrow$4&Avg \\
				\toprule[1pt]
				SASA-$\alpha$           & 83.95 & 81.92 & 79.85 & 82.96 & 83.94 & 80.99 & 77.05 & 80.11 & 78.33 & 68.65 & 70.62 & 72.23 & 78.38 \\
				SASA-$\beta$           & 83.93 & 81.47 & 78.82 & 81.88 & 83.70 & 80.98 & 76.73 & 79.75 & 77.98 & 68.23 & 70.35 & 74.98 & 78.23 \\
				SASA-IV-$\alpha$           & 84.44 & 84.96 & 80.10 & 81.59 & 82.66 & 81.08 & 77.85 & 79.66 & 78.18 & 73.99 & 77.40 & 77.57 & 79.96 \\
				SASA-IV-$\beta$           & 84.96 & 85.27 & 80.29 & 82.75 & 83.13 & 81.63 & 77.27 & 79.18 & 78.14 & 73.90 & 76.60 & 77.15 & 80.02 \\
				SASA-IV-$\gamma$           & 84.95 & 85.63 & 80.52 & 82.72 & 84.18 & 81.27 & 77.57 & 79.93 & 78.02 & 74.28 & 76.73 & 76.22 & 80.17 \\
				SASA-IV-$C$           & 84.91 & 84.93& 80.67 & 82.65 & 84.73 & 81.47 & 78.08 & 80.35 & 78.64 & 74.58 & 78.00 & 77.92 & 80.58 \\
				SASA-IV                & \textbf{85.80}     & \textbf{86.25}     & \textbf{81.32} & \textbf{83.48} & \textbf{84.74} & \textbf{82.42} & \textbf{78.67} & \textbf{80.43} & \textbf{79.04} & \textbf{74.86} & \textbf{78.58} & \textbf{79.03} & \textbf{81.22} \\
				\toprule[1pt]
		\end{tabular}}
		
		\label{tab:mimic3_ablation}
	\end{table*}

\begin{table}[t]

		\caption{AUC score(\%) on boiler fault detection for ablation study.}
		\centering
		\begin{tabular}{p{1.7cm}|p{0.55cm}p{0.55cm}p{0.55cm}p{0.55cm}p{0.55cm}p{0.55cm}p{0.55cm}p{0.55cm}}
			\toprule[1pt]
			\small{Method}&\small{1$\rightarrow$2}&\small{1$\rightarrow$3}&\small{3$\rightarrow$1}&\small{3$\rightarrow$2}&\small{2$\rightarrow$1}&\small{2$\rightarrow$3}&\small{Avg} \\
			\toprule[1pt]
            \small{SASA-$\alpha$ }                 & \small{70.62} & \small{95.19} & \small{92.59} & \small{59.79} & \small{87.81} & \small{92.98} & \small{83.16} \\
			\small{SASA-$\beta$ }                 & \small{70.22} & \small{94.87} & \small{93.01} & \small{60.07} & \small{87.44} & \small{92.77} & \small{83.06} \\
			\small{SASA-IV-$\alpha$ }                 & \small{71.97} & \small{95.07} & \small{93.26} & \small{61.68} & \small{91.48} & \small{93.14} & \small{84.43} \\
			\small{SASA-IV-$\beta$ }                 & \small{70.43} & \small{94.71} & \small{93.00} & \small{61.74} & \small{91.49} & \small{93.58} & \small{84.16} \\
			\small{SASA-IV-$\gamma$ }                 & \small{72.64} & \small{95.45} & \small{93.15} & \small{61.98} & \small{91.76} & \small{93.88} & \small{84.81} \\
			\small{SASA-IV-$C$}                 & \small{72.05} & \small{94.87} & \small{92.78} & \small{63.69} & \small{91.41} & \small{93.63} & \small{84.74} \\
			\small{SASA-IV}                 & \small{\textbf{73.61}} & \small{\textbf{95.64}} & \small{\textbf{93.77}} & \small{\textbf{65.28}} & \small{\textbf{92.37}} & \small{\textbf{94.75}} & \small{\textbf{85.90}}\\
			\toprule[1pt]
		\end{tabular}
		
		\label{tab:boiler_ablation}
	\end{table}

\subsubsection{Results on In-hospital Mortality Prediction Dataset.}
Finally, we also testify our method on the MIMIC-III dataset, which is chosen as the benchmark of unsupervised domain adaptation for time-series data in \cite{DBLP:conf/iclr/PurushothamCNL17}. We use the MIMIC-III dataset for mortality prediction and split the patients into 4 groups according to their ages. Unsupervised domain adaptation on mortality prediction is another practically significant task since the hospital can easily collect the records of the old patient but the data of the young are hard to access.

We choose 12 modalities described in \cite{DBLP:conf/iclr/PurushothamCNL17}. According to the experiment results shown in Table \ref{tab:mimic3}, we can learn the following lessons:
\begin{itemize}
    \item Similar to the other datasets, the SASA-IV overpasses the other comparison models on all the transfer tasks. Some domain adaptation tasks such as $1 \rightarrow 4$ and $2 \rightarrow 4$ even achieve 4.02 and 6.54 improvement respectively.
    \item We also find that the transfer task with large age pairs like $1 \rightarrow 4$ and $3 \rightarrow 1$ also achieve great improvement. This is because the domain discrepancy becomes larger when the age distance becomes bigger, and the proposed SASA-IV method not only aligns the domain-invariant associative structures but also considers the domain-variant information.
\end{itemize}

\subsubsection{Results on Boiler Fault Detection.}
Since the boilers are usually following stable physical rules, which are naturally considered to be transferred among different domains. Moreover, since boiler fault labeled data are very difficult to collect, it is significant to simultaneously leverage the limited labeled data and the massive unlabeled data collected from different boilers. Hence we take different boilers as different domains and consider boiler Fault detection as the time-series unsupervised domain adaptation problem.

In order to evaluate the performance of the proposed SASA-IV, we first illustrate the experimental results on the Boiler Fault Detection dataset, which are shown in Table \ref{tab:boiler}. According to the experiments, we can obtain the following observations:
\begin{itemize}
    \item Both the \text{SASA} and \text{SASA-IV} outperform the other baselines with a large margin, which proves the superior transferability. Furthermore, it is worth mentioning that the performance of the SASA-IV is much better that of SASA, which reflects the advantages of the improved strategy of sparse associative structure alignment.
    \item We also find that the other latest baselines like AdvSKM and CODAT also perform better than the baselines in the conference version like VRADA. 
    \item Similar to the SASA in conference version, SASA-IV promotes the AUC score substantially on tasks, e.g. $1\rightarrow 2$ and $3\rightarrow 2$, which are respectively improved by 4.99 and 6.9. On other easy tasks like $1\rightarrow 3$ and $2\rightarrow 3$, our method still achieves comparable results.
\end{itemize}

\subsection{Ablation Study and Visualization}

\subsubsection{The study of the effectiveness of different model variant} 
In this subsection, we provide the results of different model variants to verify the effectiveness of our method.
The experiment results on each dataset are respectively shown in Table \ref{tab:air_quality_ablation}, \ref{tab:mimic3_ablation}, \ref{tab:boiler_ablation}. According to these experiment results, we can learn the following lessons:
\begin{itemize}
    \item Compared with SASA-IV and SASA-IV-$\alpha$, we can find that the performance of SASA-IV-$\alpha$ drops, this is because $\mathcal{L}_\alpha$ can restrict the common segment length, so the sideeffect of different time lags will be removed.
    \item Compared with SASA-IV and SASA-IV-$\beta$, we can also find that the performance of SASA-IV-$\beta$ drops. These experiment results indirectly reflect that the sparse associative structures vary with different domains and aligning the sparse associative structures can avoid the sideeffect of domain-specific association. We also find that SASA-IV-$\beta$ still performs better than most of the baselines, reflecting that the extracted sparse associative structures can remove most of the redundant relationships and make the model robust.
    \item we also explore the effectiveness of domain-variant information. Compared with SASA-IV and SASA-IV-$\gamma$, we can find that the performance of SASA-IV-$\gamma$ degenerates. For one thing, these experiment results indirectly show that the strengths of associative structures vary with different domains and for another, our SASA-IV can extract and leverage the domain-variant information to achieve more robust performance. 
    \item In order to evaluate the effectiveness of the proposed unidirectional alignment restriction, we devise the SASA-IV-$\mathcal{C}$. According to the experiment results of SASA-IV-$\mathcal{C}$ in different datasets, we can find that: 1) The performance of SASA-IV-$\mathcal{C}$ is comparable with that of SASA on the air quality forecast datasets and better than SASA on the other two datasets, reflecting that the strategy of seperating structures and strengths is more superior than the alignment of weighted associative structures. 2) Compared with SASA-IV and SASA-IV-$\mathcal{C}$, we can find that the performance of SASA-IV-$\mathcal{C}$ is lower than that of SASA-IV, this circumstance indirectly proves that the bidrectional alignment will result in the wrong associative structure discovery, which further leads to the suboptimal performance. With the help of unidirectional alignment restriction, we can address this issue.
    \item We also consider the ablation experiments of SASA. Compared the result of SASA and SASA-$\alpha$, we can find that the performance of SASA-$\alpha$ drops. This is because of $\alpha$ represents the probability of the length of a segment. And the duration of segments varies with different domains. With the restriction of $\alpha$, we can exclude the influence of domain-specific segments duration.
    \item According to the experiment results of SASA-$\beta$, we can find that the performance of SASA-$\beta$ is worse that the standard SASA. This is because the sparse associative structure have been extracted, which is also more robust than that of normal feature extractor. But the reserved domain-specific associative relationships lead to the suboptimal results. Note that the SASA-$\beta$ is still better than the baselines shown in Table \ref{tab:air_quality}, \ref{tab:mimic3} and \ref{tab:boiler}. This is because the $\mathcal{L}_\alpha$ aligns the offsets between different domains, which benefits to extracting sparse associative structure for adaptation.
\end{itemize}

\subsubsection{Visualization of Aligned Associative Structures.}
\begin{figure*}[hbpt]
	\centering
	\subfigure[Heatmap of SASA]{
	\begin{minipage}[t]{0.97\columnwidth}
	\includegraphics[width=\columnwidth]{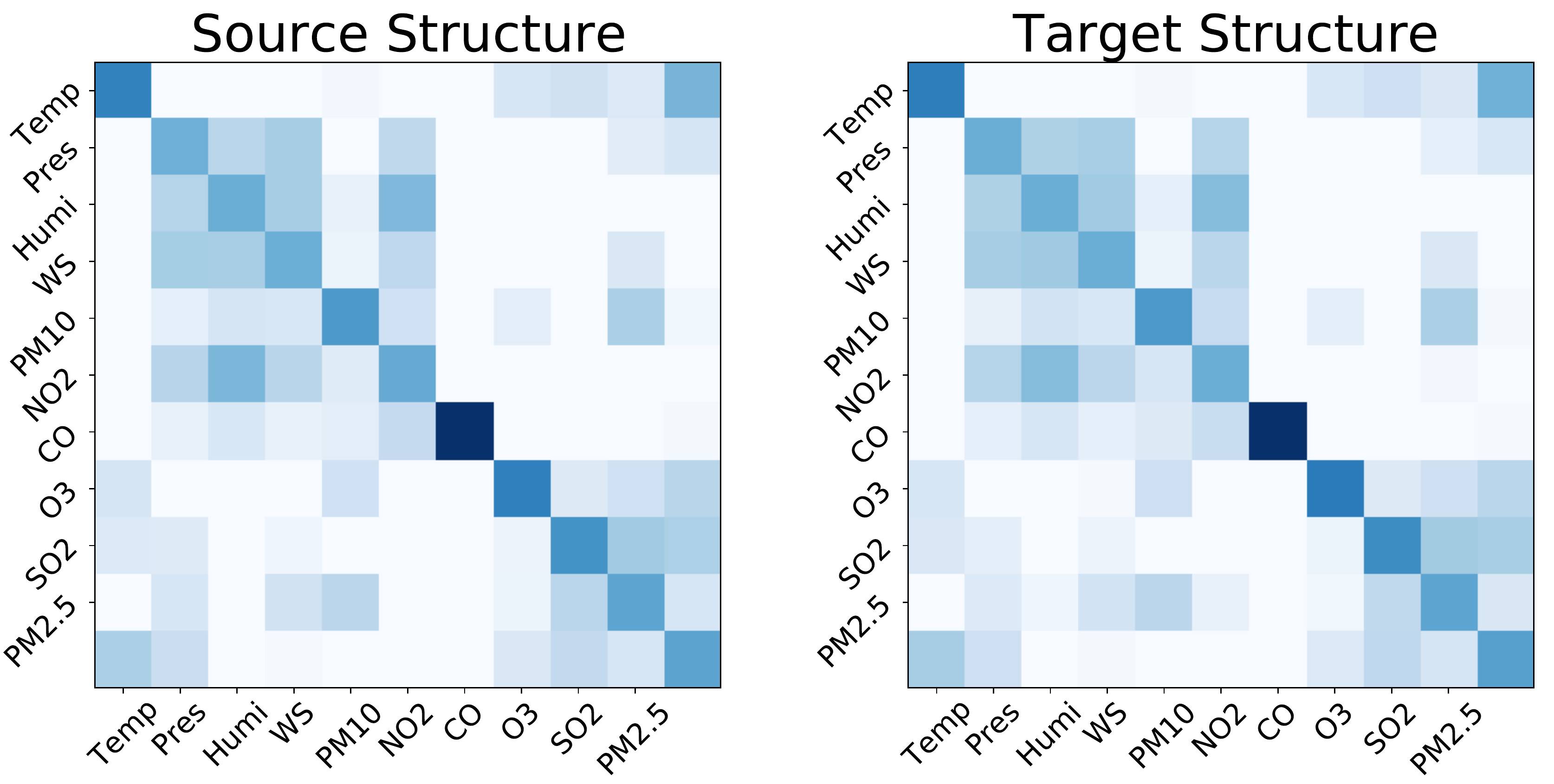}
	\end{minipage}
	}
	\subfigure[[Heatmap of SASA-IV ($\epsilon$=0.0)]{
	\begin{minipage}[t]{0.97\columnwidth}
	\includegraphics[width=\columnwidth]{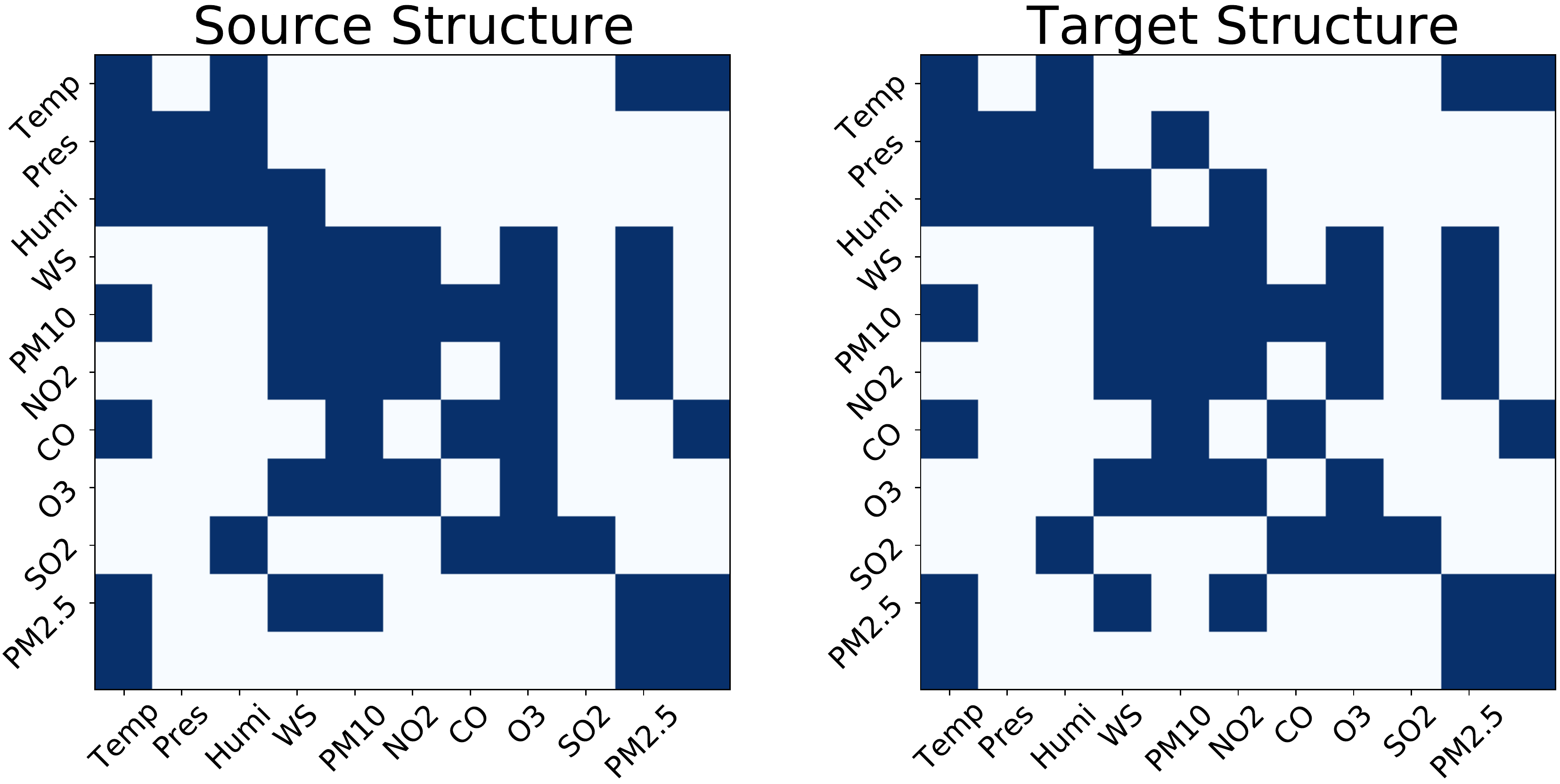}
	\end{minipage}
	}
	\subfigure[[Heatmap of SASA-IV ($\epsilon$=0.1)]{
	\begin{minipage}[t]{0.97\columnwidth}
	\includegraphics[width=\columnwidth]{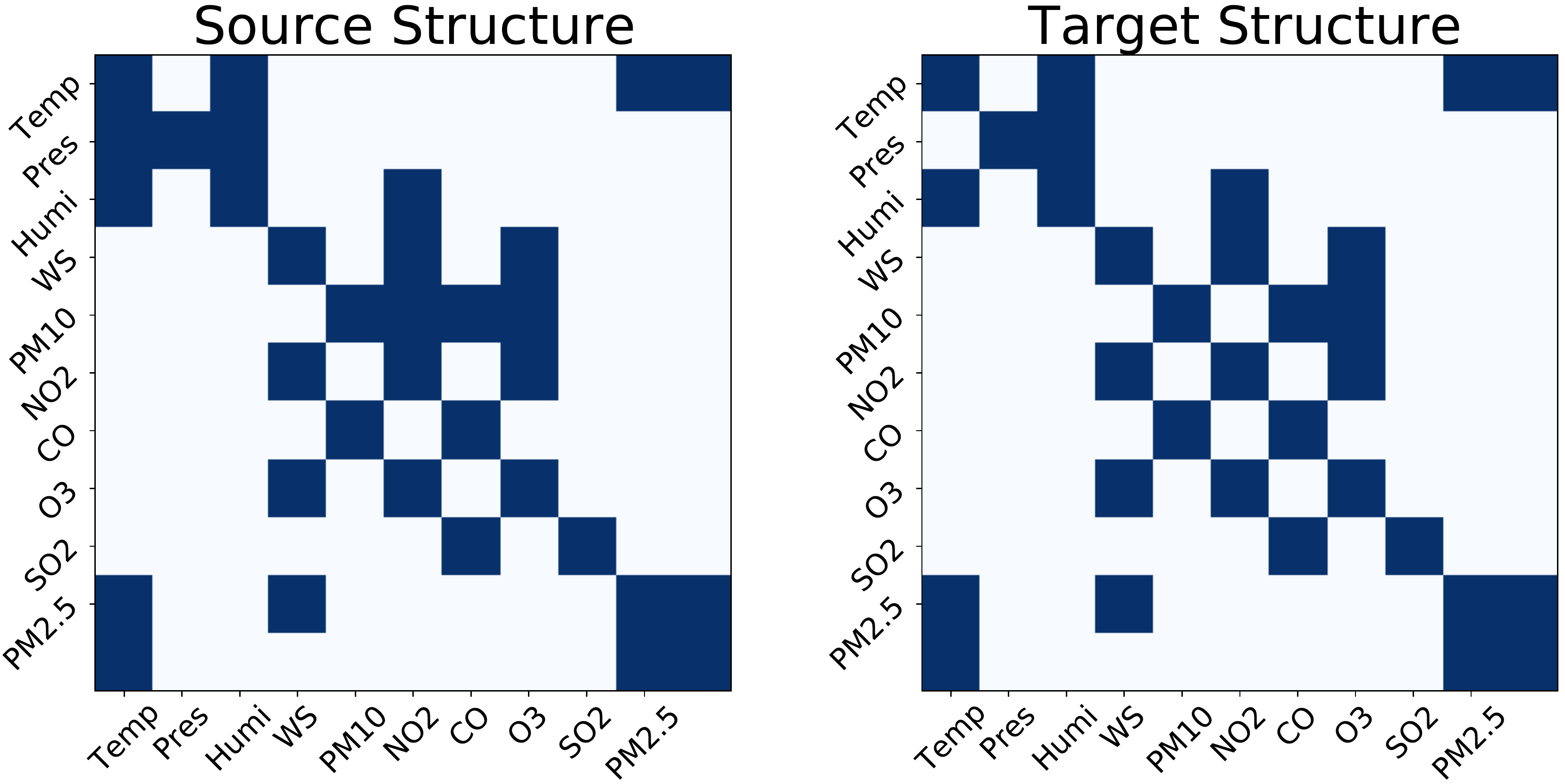}
	\end{minipage}
	}
	\subfigure[[Heatmap of SASA-IV ($\epsilon$=0.3)]{
	\begin{minipage}[t]{0.97\columnwidth}
	\includegraphics[width=\columnwidth]{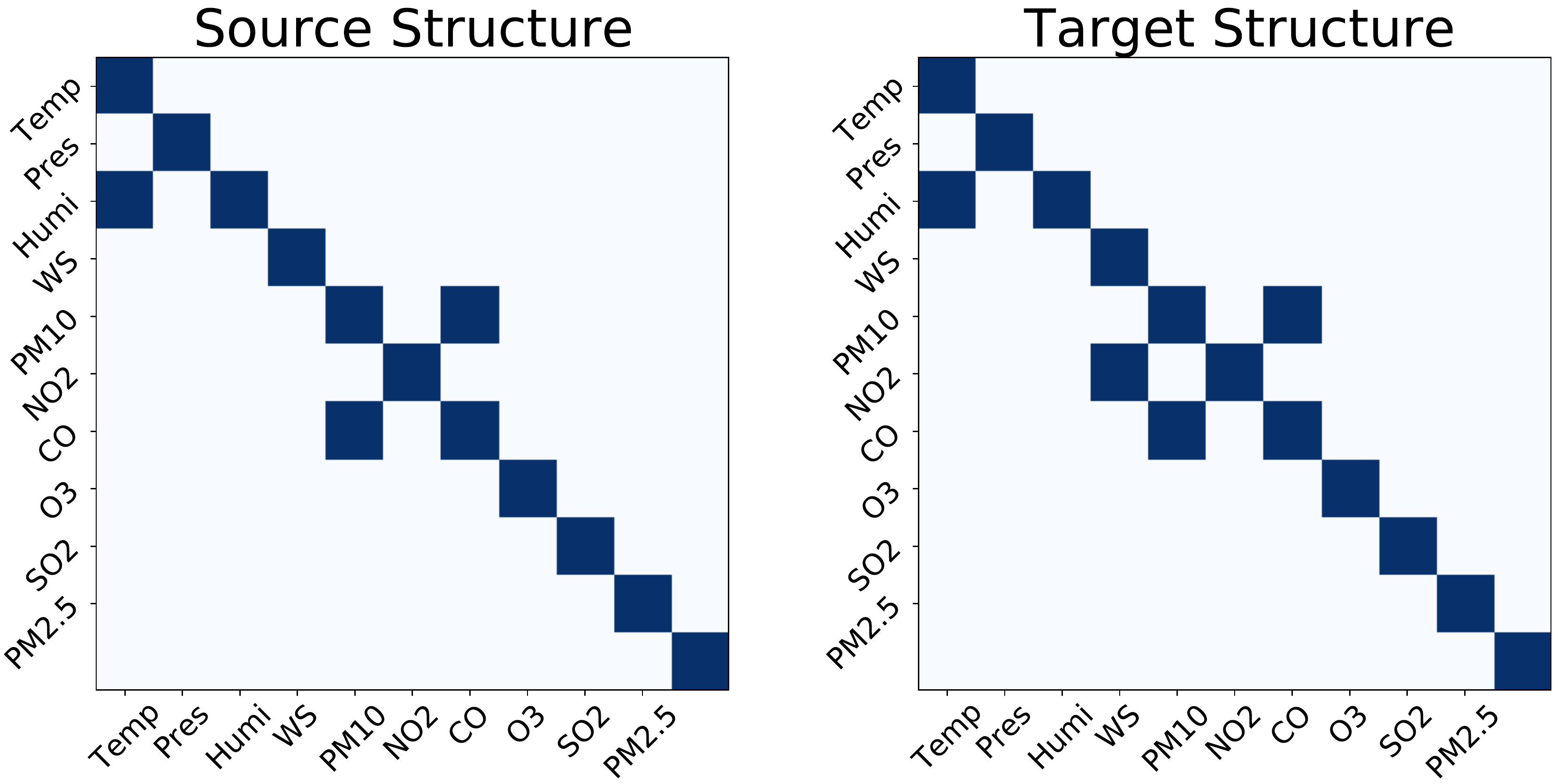}
	\end{minipage}
	}
    \caption{The illustration of visualization of correlation structure adjacent matrices under different values of $\epsilon$.}
    \label{fig:epo}
\end{figure*}

The SASA in the conference version simply leverages the Sparsemax and automatically generates the sparse associative structures. However, some redundant relationships might remain. In order to address the aforementioned issue, the proposed SASA-IV controls the sparseness of the learned associative structures by controlling the value of $\epsilon$, which is shown in Figure \ref{fig:epo}. The visualization shows that:
\begin{itemize}
    \item Figure \ref{fig:epo}(a) illustrates the heatmap of weighted associative structure of SASA. Deeper the color is , the stronger the relationships is. We can find that the color shades of the heatmap from different domain are almost the same, showing that the original SASA discard the domain-variant strengths. 
    \item The structures from different domains have many shared associative relationships, which reflects the domain-invariant mechanisms.
    \item The associative structures are very sparse, even when $\epsilon=0$, reflects that our methods can extract the sparse associative structures.
    \item The larger $\epsilon$ is, the more sparse the associative structures are. Hence our method can control the sparseness of the associative structures and remove more redundant relationships.
\end{itemize}

\section{Conclusion}\label{sec:con}
This paper presents an improved sparse associative structure alignment model for time-series unsupervised domain adaptation. In our proposal, we explore what is invariant and variant in time-series data and provide insights into how to devise the model for time-series unsupervised domain adaptation. Technically, the weights for unweighted sparse associative structure alignment are embedded and the gradient stopping is employed for better common structure discovery. We further take domain-variance information into consideration with the help of autoregressive feature extraction. The success of the proposed sparse associative structure alignment method not only provides an effective and novel solution for time-series domain adaptation but also provides some insightful theorems and results on what transfer to learn and how to achieve the ideal transfer.


%

\ifCLASSOPTIONcompsoc
  \section*{Acknowledgments}
\else
  \section*{Acknowledgment}
\fi

The authors would like to thank Zhenjie Zhang and Xiaoyan Yang from the PVoice Technology as well as Zhuozhang Li from the Guangdong University of Technology for their help and support on this work.



\ifCLASSOPTIONcaptionsoff
  \newpage
\fi



\bibliographystyle{IEEEtran}
\bibliography{bare_jrnl_new_sample4}
\end{document}